# Generating an interactive online map of future sea level rise along the North Shore of Vancouver: methods and insights on enabling geovisualisation for coastal communities

Forrest DiPaola
0009-0007-1607-4038
forrestdip@gmail.com

Anshuman Bhardwaj
0000-0002-2502-6384
anshuman.bhardwaj@abdn.ac.uk

Lydia Sam
0000-0003-3181-2960
lydia.sam@abdn.ac.uk

School of Geosciences, University of Aberdeen, Aberdeen, UK

## Abstract

Contemporary sea level rise (SLR) research seldom considers enabling effective geovisualisation for the communities. This lack of knowledge transfer impedes raising awareness on climate change and its impacts. The goal of this study is to produce an online SLR map accessible to the public that allows them to interact with evolving high-resolution geospatial data and techniques. The study area was the North Shore of Vancouver, British Columbia, Canada. While typically coarser resolution (10m+/pixel) Digital Elevation Models have been used by previous studies, we explored an open access airborne 1 metre LiDAR which has a higher resolution and vertical accuracy and can penetrate tree cover at a higher degree than most satellite imagery. A bathtub method model with hydrologic connectivity was used to delineate the inundation zones for various SLR scenarios which allows for a not overly complex model and process using standard tools such as ArcGIS and QGIS with similar levels of accuracy as more complex models, especially with the high-resolution data. Deep Learning and 3D visualizations were used to create past, present, and modelled future Land Use/Land Cover and 3D flyovers. Analysis of the possible impacts of 1m, 2m, 3m, and 4m SLR over the unique coastline, terrain and land use was detailed. The generated interactive online map helps local communities visualise and understand the future of their coastlines. We have provided a detailed methodology and the methods and results are easily reproducible for other regions. Such initiatives can help popularise community-focused geovisualisation to raise awareness about SLR.

## 1. Introduction

During the last seventy years the population of coastal cities has expanded 4.5 times (Xu et al., 2021). 230 million people live within less than 1 metre (1m) above an open body of water whereas 1 billion live less than 10m above (Kulp and Strauss, 2019). It is likely that hundreds of millions of people will be displaced during the next centuries due to Sea Level Rise (SLR) (Murali and Kumar, 2015; Mahapatra et al., 2015). From 1901 to 2018 there was between 15-25cm increase in average sea level and 7.5cm of that increase occurred from 1993 to 2017 (Frederikse et al., 2020). Due to the number of people that will be affected, there have been numerous studies throughout the world on how SLR will impact various communities. Most of this research has been completed at a local level (Rwanga and Ndambuki, 2017). However, when it comes to developing effective geovisualisation of the SLR research results for coastal communities and policy makers to realise the seriousness of the situation, many efforts are still needed.

    Numerous studies have predicted that over the coming years an increase of SLR will occur because of the global temperature increase due to Climate Change, the extent of that increase varies greatly (Sahin et al., 2019). The Intergovernmental Panel on Climate Change (IPCC) predicts that at a Representative Concentration Pathway (RCP) of 2.6 (one of the best-case scenarios) there will likely be a global mean SLR of 0.30-0.65m by 2100 and 0.54-2.15m by 2300. At RCP 8.5 (one of the worst case scenarios) there is likely to be an increase of 0.63-1.32m by 2100 and 1.67-5.61m at 2300 (Horton et al., 2020). Several researchers have discussed issues of properly accounting for the overall uncertainty inherent in SLR assessments (Gesch, 2018; Sirianni et al., 2012). These authors state that best practices should be used to increase the accuracy of Digital Elevation Models (DEMs) and Land Uses/Land Covers (LULCs) (the identification of major surface cover classes) in the analysis of SLR. There are many free, publicly accessible satellite global DEMs that are widely used by SLR studies including; SRTM, ASTER GDEM, ALOS World 3D, TanDEM-X, NASADEM, and MERI. Each of these have resolutions that are coarser than 10m in horizontal and their vertical accuracies can vary from several meters to tens of meters, whereas ideal SLR models require fine increments (<=1m) of data (Gesch, 2018). In the study carried out by Gesch (2018), he concluded, "higher resolution data with better vertical accuracy significantly improve assessment results." However, this is not the case for the publicly accessible satellite global DEMs mentioned above which is a significant limitation to the ability to accurately discover what areas are at risk to be inundated (Xu et al., 2021). Aforementioned global DEMs usually are also of coarser spatial resolutions, varying between 10m and 90m per pixels, thus limiting their capabilities for SLR research. Due to these spatial resolution and accuracy limitations of free-to-access global DEMs, there is a growing need to employ higher resolution and better accuracy regional-scale DEMs to enable reliable SLR estimates. In this study, we have used a high resolution (1 m/pixel) and high accuracy publicly available airborne Light Detection and Ranging (LiDAR) DEM. Such LiDAR DEMs are usually better than 3m/pixel in spatial resolutions (Rwanga and Ndambuki, 2017; Breili et al., 2020). Besides resolution, LiDAR DEMs have high vertical accuracy which is very important in estimating land area and property vulnerability to SLR (Sirianni et al., 2012). Lastly, LiDAR is better at penetrating the vegetation canopy which creates more reliable DEMs (Vernimmen et al., 2020), a valuable improvements for deriving the terrain of vegetated coasts.

    There are many types of models of SLR predictors including Coastal Impact Visualization Environment (CIVE), Coastal Risk Assessment Frame (CRAF), Dynamic Interactive Vulnerability Assessment (DIVA) and Bathtub Model. CIVE and CRAF are both models that are used for specific regions in the USA and the UK. DIVA that is very expensive to use requires a significant amount of data to create simulations such as tidal patterns which are not part of the scope of

this paper. By contrast, the bathtub model is one of the most popular simulations for predicting SLR due to its quick development and its ease of use for the public to understand. This model uses simple mass balance equations based on the premise of water entering a tub. From this concept a relatively reliable model of SLR can be created in a short period of time. Moreover, when Hydrologic connectivity (inundation only occurs in areas that are directly connected to the ocean (Fu and Song, 2017)) is added to the bathtub method, it can perform at levels close to the accuracy of more complex models. As GIS data collection advances, an expanding volume of publicly accessible LiDAR mapping data will become more available.

Our study area, the coastline of Vancouver, is predicted to display 1.0-1.4m SLR by 2100, and 2m SLR by 2200 on the website of the City of Vancouver at: [https://vancouver.ca/green-vancouver/climate-change-and-sea-level-rise.aspx#:~:text=Based%20on%20sea%20level%20rise,metres%20(6%20feet)%20by%202200](https://vancouver.ca/green-vancouver/climate-change-and-sea-level-rise.aspx#:~:text=Based%20on%20sea%20level%20rise,metres%20(6%20feet)%20by%202200) (Vancouver, 2023). A study (Malik and Abdalla, 2016) predicted the worst case of SLR would be 4m for the region by 2300. However, it is worth noting that such estimates are still based on lesser-to-moderate impact scenarios, and any unforeseen or extreme climate scenario can lead to far-reaching impacts on this coastline. Utilising the free-of-cost availability of LiDAR data and computationally efficient algorithms for SLR predictions, thus becomes relevant to generate more reliable models for this coastline, which can further be visualised and interacted with by the local communities.

Geovisualisation enables visual analysis of geospatial data and is achieved through convergence of cartography, geographic information system (GIS), and geomodelling methods (Kraak and Ormeling, 2010). In contemporary geovisualisation, digital maps are the base datasets, and improving internet connectivity and online map hosting platforms are further aiding to increasing number of researchers thinking in this direction. Coastal communities are one of the most vulnerable to natural disasters, and SLR and associated impacts have the potential to alter community socioeconomy drastically. Thus, a focus of SLR research should also be on designing and advancing efforts to enable digital geovisualisation of findings to the community and local policymakers. With free-to-access geovisualisation platforms such as Environmental Systems Research Institute (ESRI) Story Map becoming popular, now is the time to ensure that complex SLR results can be made accessible online in an easy to understand format for the public. Even policymakers are often not from geophysical backgrounds and easily understandable spatial geovisualisation can help their understanding for quick and informed decision making. However, coastal and SLR geovisualisation is still in its initial years with only several significant publications and that too mostly post-2015 (e.g., Minano et al., 2018; Newell and Canessa, 2017; Yulfa et al., 2018). Thus, there is a need to form a framework that implements geovisualisation on various SLR scenario over a platform that can be made easily accessible to public.

Keeping in view, the aforementioned research gaps and needs, the aim of our paper is to create a more precise, publicly accessible interactive online map using modern GIS methods that allows the public to explore the impacts of SLR due to Climate Change with its direct impact to key well-known places of the North Shore of Vancouver, Canada using new high-resolution data and Deep Learning.

To achieve this aim, there are three main objectives:

1. Demonstrate which areas will be inundated on the coast at 1m, 2m, 3m, and 4m SLR by using 1m resolution LiDAR Digital Elevation Model (DEM) implementing the bathtub method model with hydrologic connectivity.
2. Create a land use map of these areas and predict future changes in land use with modern methods including Deep Learning.
3. Create an interactive online map detailing findings for the public to explore negative SLR impacts to local major infrastructure, commercial areas, and residential areas.

## 2. Study Area

The study area is the North Shore of Vancouver (Figure 1), British Columbia, Canada. This region is surrounded by three bodies of water: Howe Sound to the west, Burrard Inlet to the south and Indian Arm to the east. The northern third of the District of North Vancouver and West Vancouver was left out of this study since most of the inhabited areas of this mountainous area are ski resorts which will not be affected by SLR.

The Greater Vancouver area is considered to be amongst the most vulnerable cities to SLR (Lyle and Mills, 2016). Compared to the rest of the Metro Vancouver area, the North Shore relies heavily on its coastal economy and includes part of the largest commercial port in Canada. Unlike the City of Vancouver (Lyle and Mills, 2016) and some of its suburbs such as Richmond (Malik and Abdalla, 2016), there have been no major focused studies completed on the effects of SLR on the North Shore. The temperate climate of the Greater Vancouver area is oceanic, humid, and cool with a significant rainy season normally lasting from October to March. The average temperature is 11.0°C. This region is susceptible to flooding from the Fraser River as well as to earthquakes and windstorms.

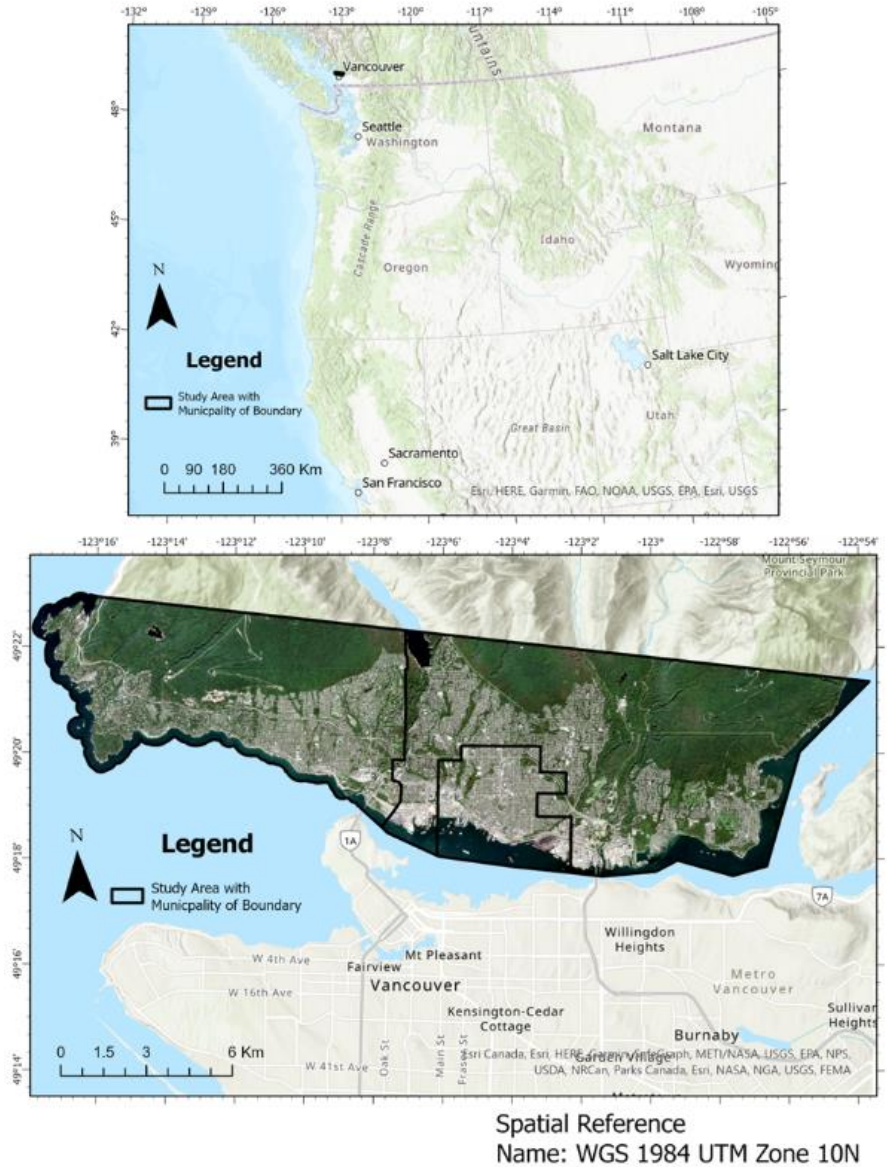

Figure 1 Maps of the overall location of the study area (top) and the local area with boundaries of the Municipalities (bottom)

# 3. Methodology

## 3.1. Creating SLR images

Unlike most DEMs used in the past for SLR created from typically 10m+ satellite images, this project used airborne LiDAR that is 1m in resolution. LiDAR data has a higher resolution, higher vertical accuracy, and can penetrate tree cover at a higher degree than most satellite imagery (Sirianni et al., 2012; Vernimmen et al, 2020). The bathtub method model with hydrologic connectivity was used to delineate inundation zones. This allowed for a not overly complex model and process, making it possible to obtain results by using standard tools like ArcGIS at levels close to the accuracy of more complex models especially with the high-resolution data.

LiDAR DTM dataset and the created Coastline shapefile were the two layers used to create the four SLR levels. High resolution LiDAR DTM also allowed the creation of a more accurate coastline with better vertical accuracy. A polygon was created by hand of the high tide coastline of the study area which was used for a mask throughout this Methodology.

The Raster Calculator was an important tool to create a new raster image to show 0 to 1m of SLR using the equation "DEM" <=1. This demonstrated all the areas that would be inundated if the sea level rose to 1m. 1m was used since the City of Vancouver predicts that there will be 1.0-1.4m SLR in the region at the current trajectory for global temperature rise by 2100. Moreover, 1m is the smallest increment in the LiDAR DEM. The above was repeated for 2m, 3m and 4m.

Since the four equations that were used produced all areas between 0m and 4m above sea level, several of the areas that were within this range were located inland without any connection to the coastline of Burrard Inlet. To rectify this, all areas that did not connect to the inlet were deleted by using hydrologic connectivity, accomplished by converting the raster image to a vector format. Even though there might be some degradation of data, whenever converting between the two, vector data was needed for the land use map. After the four new polygon layers (1m, 2m, 3m, and 4m) were created, removing inland inundation zones was possible. This was accomplished by using the four polygon layers with the Coastline shapefile and using Select by Location with Are Crossed by the Outline of the Source Layer Feature as the selection method. This allowed for areas that touch the borders between 0 and 4m to be the only areas. This approach of hydrologic connectivity worked for this study area since three of the four sides of the borders are water bodies and the last is a mountainous area with very high elevation. These new polygons were the end product for the four SLR layers used throughout the next sections of the Methodology.

## 3.2. Creating past, present, and future Land Use

A 2021 LULC, part of the Sentinel-2 10m Land Use/Land Cover Time Series, was taken from Esri Living Atlas. The LULC was clipped with the Coastline shapefile that was already created. This allowed for a similar study area of the region. However, since this raster had a resolution of 10m and the SLR data had one of 1m, the clipping and results were less accurate. The four SLR layers clipped the output of the LULC to demonstrate four different LULCs of the four different heights of SLR for analysis to examine what landcover would be inundated at each SLR height.

To test the accuracy of ESRI's 2021 LULC, an accuracy assessment was completed by using ArcGIS Pro and Google Earth Images. In ArcGIS Pro, 500 stratified random points were used to observe if the classification was the same as the ground truthing. For this LULC ground truthing was completed by carrying out a small number of field observations, coupled with Google Earth Images for the year of 2021. From this a Confusion Matrix was created which demonstrates a high overall accuracy of 97% and a high Kappa Coefficient of 94% for the 2021 LULC.

Since there will be changes for the current LULC of the study area when 1m to 4m SLR might occur, this study used past land cover changes to predict future ones in TerrSet Idrissi. This was particularly important because of the long-term time range to achieve 1m to 4m rises in sea level, if at all. In TerrSet LULC maps were created from Landsat 5 images from the years 1991, 2006 and 2011. By using Land Change Modeler of this program with the first two land cover maps as well as other maps such as distance from roads, the modeler loosely predicted future LULC. The 2011 image was used for validating the results from the MLP/MC Deep Learning prediction that was created by the first two images of 1991 and 2006.

Landsat 5 images from 1991, 2006 and 2011 were used. All the satellite images that were downloaded were from Tier 1, the highest quality of data. Moreover, Level 2 data was used since Bottom of Atmosphere corrects atmospheric effects resulting in surface reflectance images. These satellite images were taken over the summer period to avoid the presence of snow in the data. The Bands 2 (green), 3 (red), and 4 (near infrared) were used for the three time periods and created into composite images. A false colour image was created with Band 2 used as the blue image, Band 3 as the green, and Band 4 as the red. False colour image was used since it allows for easier training to distinguish the difference between Grassland (light red), Woodland (dark red), Bare Earth (white to very light blue) and Urban Areas (light blue).

From a supervised classification (Maximum Likelihood) of the data, five different classes were sampled to be trained and saved for each of the composite images of the three time periods. 10% of each image was trained for the remaining 90%.

The five classes were named: 1-Waterbodies, 2-Trees, 3-Grassland, 4-Buildings or Roads, and 5-Bare Earth. The bands used to create the composites for each year were then used to allow the system to analyse the sampled data. These five classes were used since each of these was defined for a program to train, and each was differently affected by SLR. Using Maximum Likelihood Classification tool (MAXLIKE), a landcover map was created for these three time periods by classifying the images with the samples. MAXLIKE was used for this study since it is one of the most popular methods of classification in remote sensing that predicts to which class each pixel belongs by the closest indicator.

To assert the accuracy of these three LULCs, three other accuracy assessments were completed on ArcGIS Pro of the respective years. However, unlike the assessment of the 2021 LULC, only Google Earth Images for each of the past years were used as the ground truths equivalent. Since there were no Google Images for 1991, the visual analysis of satellite image used to create the LULCs was the basis of the ground truth validations for that year. Each of the years had a high overall accuracy (96% for 1991, 99% for 2006, and 97 for 2011) and a high Kappa Coefficient (95% for 1991, 98% for 2006, and 94 for 2011). The lowest user accuracy was Grassland for 1991 and Bare Earth for 2006 and 2011. The reason Grassland might have been misclassified the most for the 1991 LULC is due to only having a 30-metre resolution satellite image as a reference to conduct ground truthing.

From these 3 LULCs, the Land Change Modeler of TerrSet was used to predict future change of land cover for the study area. The 1991 and 2006 LULCs were the two inputs for Land Change

Modeler from which TerrSet ran a change analysis of the gains and losses of the classes by pixel for the two time periods as well as of spatial trends of change. These will be further discussed in the Results section. Since twenty transitions can occur between the five classes, only transitions of 5000 pixels or greater were used for this study. This high threshold for transition area was used since it only allowed for one transition to occur which was Grassland to Buildings or Roads. If a lower threshold were used, then the accuracy for this Land Change Model would be less than 50%. This is discussed more in the Results section. Six variables in the form of rasters were created to be used for the MLP AI technique Sub-Model for predicting transitions of the LULC classes. These six rasters were Elevation Distance from Roads, Slope, Distance from Disturbance, Distance from Urban, and Distance from Rivers. These six variables were used with the 1991 and 2006 LULCs to predict how transitions would occur in this study area by creating Transition Potential with MLP. Once this was created, it allowed for the prediction of land change for future dates with the machine learning tool Markov Chain. For this study four dates were created which were 2011, 2100, 2200, and 2300. 2011 was used to validate the results with the formerly created 2011 LULC whereas the other three were used to predict what LULC will look like in the future.

### 3.3. Public Engagement Demonstrating Effects of Sea Level Rise

An interactive online web map was created that allows the public to view, interact, and engage with the 1m, 2m, 3m, and 4m SLR visualised output data to understand how flooding will affect their communities, including important local landmarks, sites, and infrastructure.

For the online interactive map, the same four SLR vectors were used with the additional 2021 study area LULC layer that was used using ArcGIS Online. These five layers were important for public viewing since they demonstrate which areas and land cover will be affected by SLR. A semi-transparent version of all the layers was used for this online map to allow the public to observe which areas of the satellite image will be affected such as their homes and other personal locations. The layer Municipal Boundaries was also added to the online map to show the extent of the study area.

## 4. Results

### 4.1. Results of the four study levels of SLR

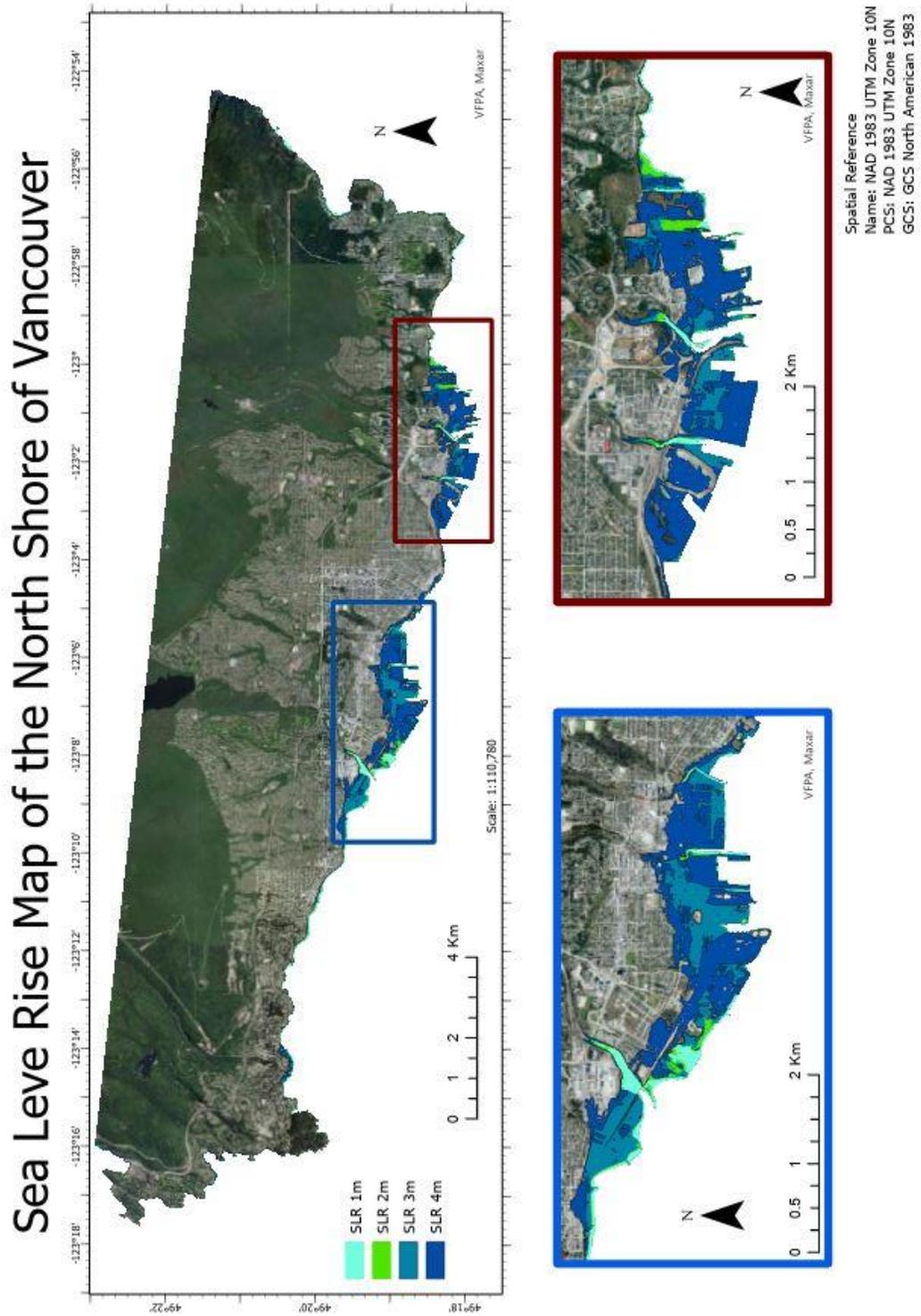

Figure 2 Possible SLR from 1m to 4m in height for the North Shore

Figure 2 demonstrates the output of the different elevations of SLR from 1m to 4m with each level of inundation being darker than the prior level. As can be seen in this map that visualizes the results of the project, all the layers combined inundate only a small swath of the study area with 1m-2m SLR flooding mainly beaches. Along the eastern and western sides of this map there is a lack of inundated area due to the rapid elevation from sea level either because of the presence of cliffs or hills adjacent to beaches.

Table 1 Inundation area and total percentage of each SLR height

| SLR height | Inundation area in $km^2$ | Percent of total study area [$150.06\ km^2$] |
|---|---|---|
| 1m | 0.78 | .5% |
| 2m | 1.09 | .7% |
| 3m | 2.93 | 2.0% |
| 4m | 6.39 | 4.3% |

Unlike the first two SLR levels, 3m SLR inundates large swaths of land that are not beaches, mostly flooding Ambleside Park in West Vancouver as well as many commercial and residential sites located along Burrard Inlet. There are also small pockets of residential areas along the coast of all three districts that become inundated.

3m to 4m increase of sea level has the greatest difference in increased area of inundation. Whereas less than 3m SLR started to inundate areas other than beaches, at less than 4m SLR most of these same areas would become completely inundated. This is the case for the North Shore Auto Mall as well as the surrounding light industrial area. Interestingly, a large portion of the area that is inundated in this level is closer to the water than the previous level, specifically regions that are near the two Narrows. This is most likely due to the raised platforms used for shipping having a higher height than the regions directly inland. There are still numerous areas that would become islands if inundation occurred at this level many of which are either railroad lines or large piles of resources such as sulphur.

### 4.2. Comparison of the Study Data to Readily Available Online Map

Figure 3 (left) is sourced from CoastalDEM, a free worldwide Interactive Sea Level Rise map created by Climate Central. Based on search results, this is one of the most widely used public maps for observing SLR globally. The horizontal accuracy for this online map for locations outside the United States (not including Alaska) is above 30 metres. Specifically for the North Shore of Vancouver, the online map has a horizontal resolution close to 30 metres throughout. CoastalDEM uses the same bathtub approach with hydrologic connectivity as this study.

Compared to the 4m that was created in this study, CoastalDEM 4m SLR seems to overpredict areas. This seems to especially be the case for areas considered to be islands in this study's data. In contrast to the results of this study, CoastalDEM predicts these islands as being inundated such as the surrounding regions. An explanation of this would be the difference in horizontal resolution between the two models. Since CoastalDEM has a pixel resolution close to thirty times larger than the model that was created for this study, areas that are distinctly either inundated or not included in this study model would be considered mixed in CoastalDEM and therefore inundated. This demonstrates the importance of having the highest horizontal accuracy for SLR since the lower the accuracy, the greater the tendency to overpredict flooding (Gesch, 2018). The image of Figure 2 is an example of the perceived overprediction and underprediction of the CoastalDEM model which predicts the whole south side to be underwater at 4m SLR and almost none of the north side inundated, and further stresses upon making SLR geovisualisation more realistic and informative for communities. By contrast the right image of Figure 3 which is the model created in this study predicts that the raised rail lines will not be flooded, whereas surrounding regions will be. The results of the regions inundated on the coast at 1m, 2m, 3m, and 4m SLR do not show a consistent increase of the areas flooded but demonstrate that with each metre of SLR the percentage of total area inundated steeply increases.

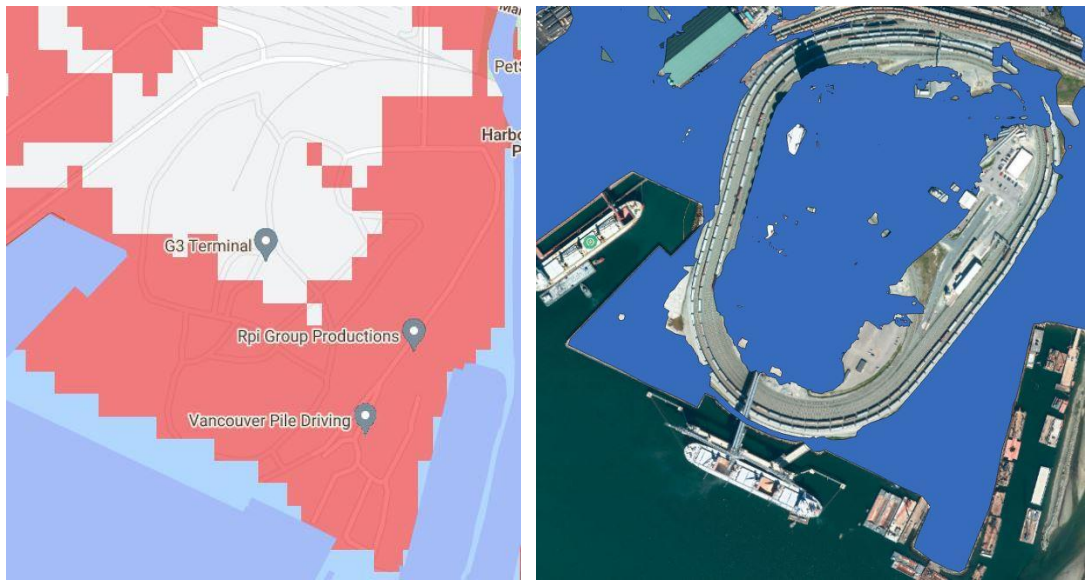

Figure 31 Left is a picture from the CoastalDEM online map; Right is an example of the same location from this study

### 4.3. Results of 2021 LULC

10N area was downloaded, masks were used on this Esri LULC that comprised each SLR layer from 1m to 4m as well as the created Coastline for the whole study area. There were five different LULC classes: Water, Trees, Grassland (which Esri names "Rangeland," but for this study was termed "Grassland" since almost all these areas are either large grass areas of parks or golf courses), Bare Earth, and Built Areas. Each of these classifications are self-explanatory besides Built Areas, which includes single houses, apartments, industrial, and commercial areas.

Table 2 demonstrates the LULC (Figure 4) classes for each metre of SLR. For 1m SLR the two most numerous classes were Built Areas and Water. Even though a mask was used of a high tide coastline, there were still some pixels for LULC that were Water from the surrounding saltwater bodies of the Burrard Inlet, Howe Sound, and Indian Arm. This is likely due to mixed pixels that were Water and another class. Moreover, instead of the resolution being 1m like the DEM that was used, the LULC used was 10m. These Water class pixels seemed also to be the mouths of rivers and streams. For 1m SLR there were barely any pixels that were the classes Grassland, Bare Earth, or Trees. With each of these levels of SLR the pixel percentage of the class increased with the exception of Water. This is due to the possibility that most of the pixels that were classified as Water were the coastline or river mouth pixels that were already calculated in 1m SLR. There was only an increase of less than 300 pixels of water from 1m to 2m SLR, which is the third lowest. Even though the least was less than 100 pixels with the class Bare Earth, the increase for this class was over 0.2%. Also, there was even more of a case for the class Grassland which increased by 1.5%. This is the opposite for Water where the class decreased by over 10% for this level of SLR. The main increase in 2m SLR was Built Areas where over 50% of the LULC was this class. Just like 2m SLR, 3m SLR has similar classification of pixel change. Built Areas increased in this level of SLR by almost 16000 pixels or an increase of over 20% which was 71.19% of all pixels being part of this class. Besides the percentage of Water decreasing, the percentage of Grassland also decreased for this level of SLR compared to 2m SLR. Most of the new area that would be inundated from 3-4m of SLR would be Built Areas. Pixels that were classified as Built Areas were 82.81% of the total area that will be inundated if 4m SLR occurred in this study area. Other than the Built Areas, each of the other classes decreased in overall percentage with both Grassland and Water decreasing by over half in their overall percentage compared to 3m SLR. Moreover, for the first time there were more pixels classified as Trees than ones classified as Water. When compared to the whole study area, there was significantly more of the class Trees compared to the areas that would be inundated. This is due to this class being mainly found at very high elevation since these regions are usually part of one of the three major mountains found in the study area.

Table 2 Number and percentage of 10m resolution pixels per LULC class for each SLR height

| SLR | Water | Trees | Built Areas | Bare Earth | Grassland | No Data |
|---|---|---|---|---|---|---|
| 1 m | 3728 at 47.94% | 203 at 2.61% | 3553 at 45.71% | 235 at 3.02% | 57 at 0.73% | |
| 2 m | 4039 at 37.13% | 735 at 6.76% | 5544 at 50.96% | 326 at 3.00% | 235 at 2.16% | |
| 3 m | 4390 at 14.97% | 2827 at 9.64% | 1079 at 71.19% | 436 at 1.49% | 596 at 2.03% | |
| 4 m | 4573 at 7.16% | 5224 at 8.18% | 52868 at 82.81% | 574 at 0.90% | 607 at 0.95% | |
| Study Area | 17818 at 1.19% | 71966 at 47.96% | 74101 at 49.38% | 6211 at 0.41% | 15866 at 1.06% | 9 at <0.01% |

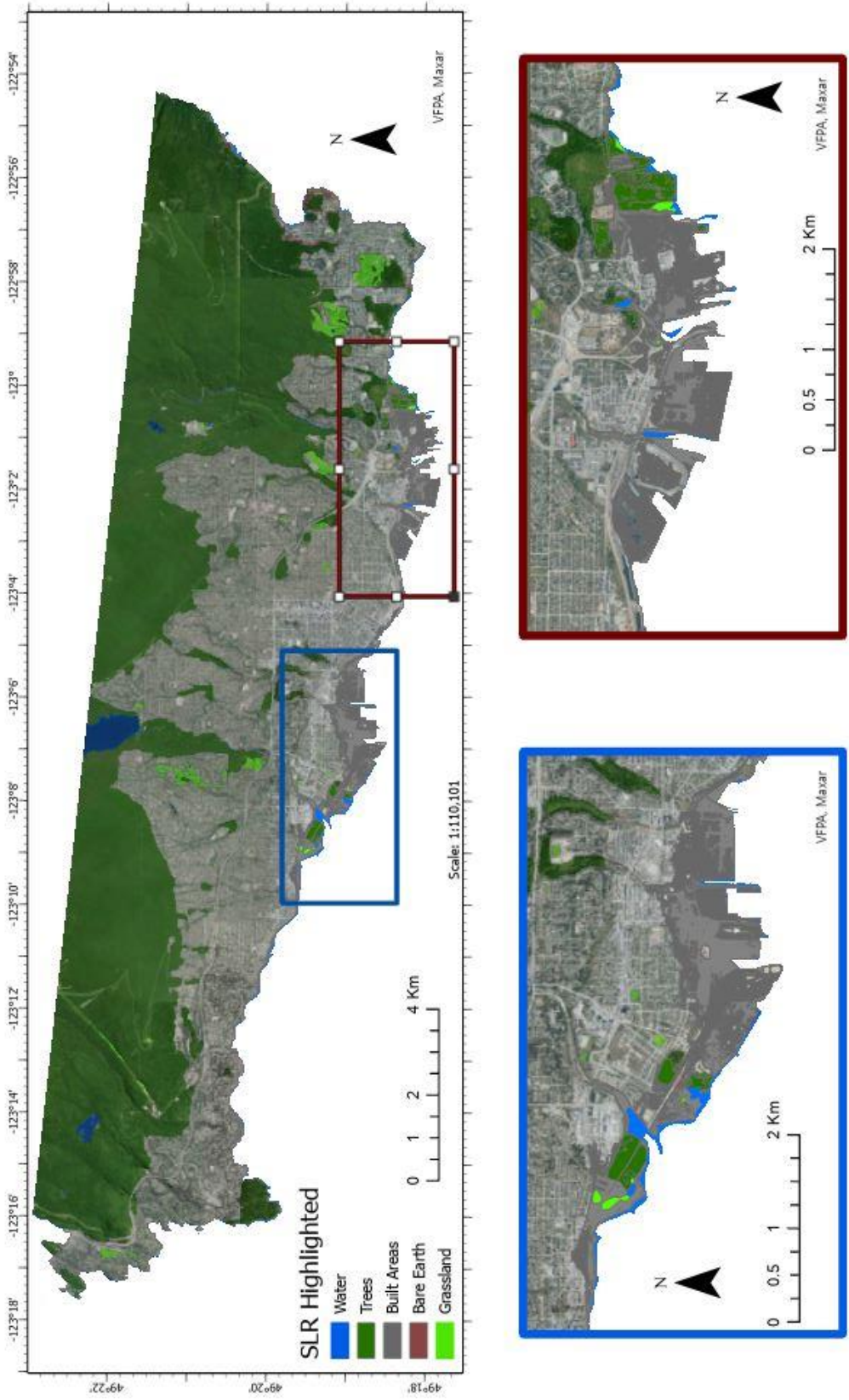

Figure 4 2021 LULC with inundated areas highlighted

## 4.4. Results from Future LULC in Inundated Areas

LULC is an important part of SLR based studies because it demonstrates what types of areas will be affected by inundation (Rwanga and Ndambuki, 2017; Lentz et al., 2019). Artificial Intelligence (AI) techniques like Deep Learning have been used in analysing remote sensing data at a rapidly increasing rate. The techniques that have been created using Deep Learning algorithms have improved the ability to categorize remote sensing images (Campos-Taberner et al., 2020). These general systems use a Deep Learning Convolutional Neural Network (CNN) approach which is the main method used in AI-based visual systems for many tasks including segmentation and identification (Solórzano et al., 2021). Land cover change is the loss of natural areas, usually forests, grasslands and other natural environments to urban development (Mihailescu and Cîmpeanu, 2020). The AI technique called Multi-Layer Perceptron (MLP) is used by Land Change Modeler to predict these future changes. MLP itself is a feedforward artificial neural network (ANN) which is a group of algorithms that try to identify relationships in particular data sets by a process that mimics the workings of the human brain (Camacho Olmedo et al., 2018). Markov Chain (MC) techniques, used to find the probability of change, is a simple model that in LULC predicts the change over time by using past trends to predict future ones. TerrSet's Land Change Modeler uses a combination of MLP and MC to quantify spatiotemporal change; MLP models the transitions whereas MC models the future predictions (Shen et al., 2020).

Since there is a likelihood of there being a change in the LULCs of the study area if SLR occurs, a Land Change Model was created to predict what the LULC would be in 2100 (Figure 7), 2200 (Figure 8), and 2300 (Figure 9). These three dates were used because the City of Vancouver predicts that around 2100 there will be between 1.0-1.4m rise, and in approximately 2200 there will be 2m rise. The date 2300 was used since the IPCC predicts that RCP of 8.5 would be between 1.67-5.61m around 2300 (Horton et al., 2020).

To predict future changes, three past LULCs were created from their satellite images (1991, 2006, and 2011) since two maps were needed to compare differences and the last map was used to validate the TerrSet AI Change Model (….). This AI Change Model allows users to train machine learning models by using time-series satellite imagery. The 1991 LULC that was created with MAXLIKE with Landsat 5 bands of 234 was masked with the shapefile Coastline to have the same area as that of the study area (150km$^2$). Trees had the largest area of 72km$^2$ which is almost the same as the 2021 LULC that was created by Esri. On the other hand, the Waterbodies class had a much smaller area in the 1991 map compared to the 2021 LULC. This is probably due to the following two reasons. The first is that the 2021 LULC used Sentinel-2 images with a resolution of 10m whereas the 1991 LULC had a resolution of 30m; rivers and other waterbodies which are quite narrow would be more likely to be identified for the 2021 LULC. Secondly, some of the surface that was in the study area was classified as sea water in the 2021 LULC whereas none was for the 1991 LULC, probably demonstrating different tidal levels between the two LULCs. Moreover, Grassland and Buildings are quite different compared to 2021 LULC with Grassland being 20km$^2$ (compared to 1.5km$^2$ in 2021) and Buildings being 55km$^2$ (compared to 74km$^2$ in 2021). Once more there are perceived to be two major differences between these two LULCs for these classes. Since there was a great deal of development in the study area during the last thirty years, it makes sense that the Buildings class would be much lower in 1991 compared to 2021. Moreover, it seems that most of the area that is part of the class Grassland in 2021 is either golf courses or parks whereas the 1991 map also includes residential lawns. This is probably due to the difference of the type of classification between the two LULCs. Bare Earth is a class that had hardly any pixels for the area with less than 0.01km$^2$ and this is similar for the 2021 LULC.

The LULCs for the other two years of 2006 and 2011 are quite similar in class distribution compared to the 1991 LULC. The only major differences are the continued increase of the class Buildings and the decrease of the class Grassland. The area of the Buildings class increased from 3km$^2$ to 58km$^2$ in 2006 and increased another 6km$^2$ to 64km$^2$ in 2011. On the other hand, class Grassland decreased by 4km$^2$ to 16km$^2$ in 2006 and to 11km$^2$ in 2011. The overall trends of the increase of Building areas and the decrease of Grassland are predictable since the most likely change would be for vegetation to become buildings due to urbanization. However, the rapid increase in five years of urbanization between 2006 to 2011 is surprising when considering that the rate of change doubled between these two dates compared to the fifteen years between 1991 to 2006. There are a few reasons that this rapid urbanization could have occurred between these dates. Firstly, there was rapid development in this location between these periods for the 2010 Winter Olympic Games including the expansion of the North Shore Highway connecting Vancouver to the Whistler ski resort. Moreover, this region has undergone densification with single house properties being replaced by condominiums or apartments creating less of an area for lawns and gardens. Lastly, there were many recent cut-blocks into forested higher elevations in the study area in 1991 classified as Grassland which in the following years were classified as Trees.

From the 1991 and 2006 LULCs and the six Structure Variables a Transition Sub-Model was created using MLP. Before creating the Sub-Model a change map was created which ignored any transitions that were less than 5000 cells. The only transition that fulfilled this requirement was the Grassland to Buildings or Roads. The reason that such a high-level of cells was used is that the MLP transition Sub-Models created would have an accuracy of less than 50% if transitions of less than 5000 cells were used.

The MLP that was created ran through 6014 samples per class and the overall accuracy of this Module was 67.60%. The overall skill measure was 0.3520 whereas the Transition: Grassland to Buildings or Roads had a skill measurement of 0.4952, and Persistence: Grassland had a measurement of 0.2087. The tabulated results are shown in Tables 4 and 5.

Out of all the variables the most influential one in this model is Distance from Roads. The overall accuracy of the model decreased by over 6% when it was removed. By itself it had an accuracy of 63.52% which was almost 7% higher than the next highest. On the other hand, Distance from Rivers was the least influential since, when it was changed to a constant, the overall accuracy only dropped by less than a percentage. When it was used as the sole variable, the module had an accuracy of 48.67%. Distance from Disturbance had the lowest accuracy when it was used as the only variable, but when it was turned into a constant, it was the third most influential variable. The difference between these two accuracy percentages could be due to Distance from Disturbance relying on other variables to function well. The other variables' percentages of accuracy had a similar mid-level ranking when changed to a constant and when being the sole variable.

Table 3 General model information - input variables

| | |
|---:|:---|
| *Independent variable 1* | Elevation |
| *Independent variable 2* | Distance from Rivers |
| *Independent variable 3* | Distance from Disturbances |
| *Independent variable 4* | Distance From Roads |
| *Independent variable 5* | Distance From Urban Areas |
| *Independent variable 6* | Slope |
| *Training site file* | 1991-2006_Train_All |

Table 4 General model information - parameters and performance

| | |
|---:|:---|
| *Input layer neurons* | 6 |
| *Hidden layer neurons* | 7 |
| *Output layer neurons* | 2 |
| *Requested samples per class* | 6014 |
| *Final learning rate* | 0.0005 |
| *Momentum factor* | 0.5 |
| *Sigmoid constant* | 1 |
| *Acceptable RMS* | 0.01 |
| *Iterations* | 10000 |
| *Training RMS* | 0.4536 |
| *Testing RMS* | 0.4565 |
| *Accuracy rate* | 67.60% |
| *Skill measure* | 0.3520 |

Table 5 Model breakdown for transition and persistence

| *Class* | *Skill measure* |
|---:|:---|
| *Transition : Grassland to Buildings or Roads* | 0.4952 |
| *Persistence : Grassland* | 0.2087 |

Table 6 Forcing a single independent variable to be a constant

| Model | Accuracy (%) | Skill measure | Influence order |
|---|---|---|---|
| With all variables | 67.60 | 0.3520 | N/A |
| Var. 1 constant | 64.87 | 0.2974 | 2 |
| Var. 2 constant | 66.73 | 0.3347 | 6 (least influential) |
| Var. 3 constant | 65.35 | 0.3071 | 3 |
| Var. 4 constant | 61.14 | 0.2229 | 1 (most influential) |
| Var. 5 constant | 66.68 | 0.3337 | 5 |
| Var. 6 constant | 66.05 | 0.3210 | 4 |

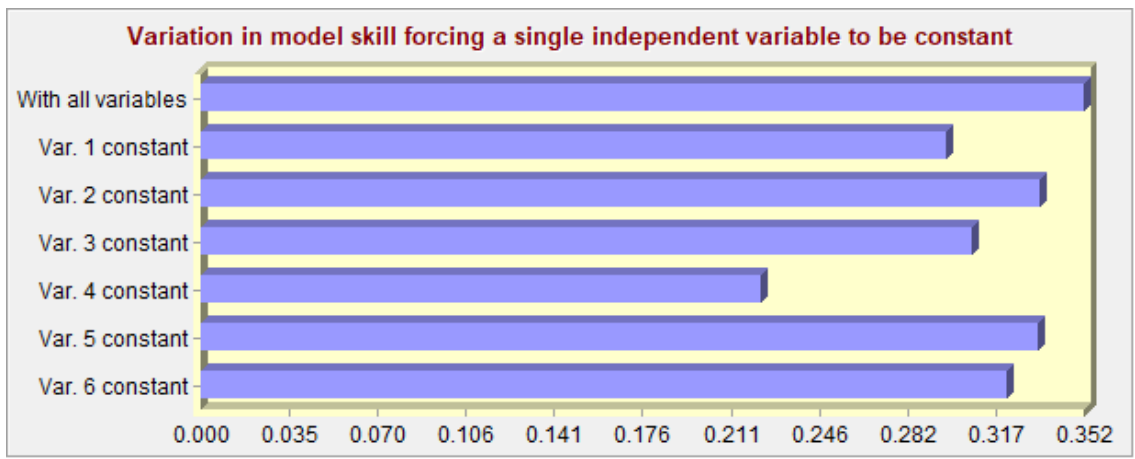

Figure 5 Graph demonstrating importance of single variable when made into a constant (smallest number most influential)

Table 7 Forcing all variables but one to be a constant

| Model | Accuracy (%) | Skill measure |
|---|---|---|
| With all variables | 67.60 | 0.3520 |
| All constant but var. 1 | 52.86 | 0.0572 |
| All constant but var. 2 | 48.67 | -0.0266 |
| All constant but var. 3 | 48.94 | -0.0213 |
| All constant but var. 4 | 63.52 | 0.2705 |
| All constant but var. 5 | 53.93 | 0.0785 |
| All constant but var. 6 | 56.05 | 0.1211 |

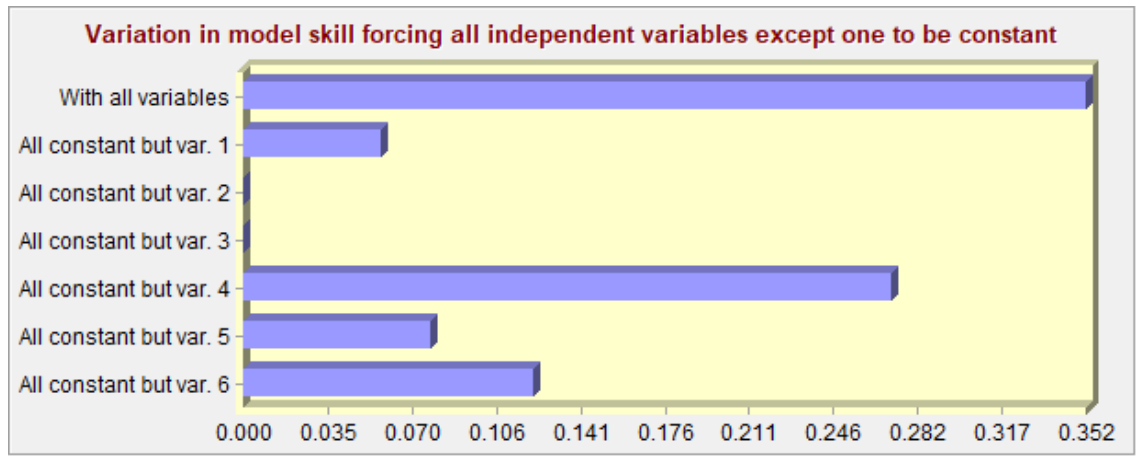

Figure 6 Graph demonstrating if only one variable is used

From this MLP a prediction LULC can be made. The first prediction created was of 2011 because this allows for the past mapped LULC of the same year to be used to validate the MLP model result. The validation that was created showed three different outcomes: hits, misses and false alarms. Hits indicates the MC predicts correctly that the cell will change; misses shows that the MC did not predict the LULC would change although a transition did occur; and, false alarms indicates that MC predicted a transition that did not occur. In general, there were not many hits that occurred throughout the predicted map vs the actual LULC map. This is probably due to there being only one transition being tested, whereas there were many actual different transitions occurring during this time period. However, when a lower accuracy model was created with six transitions, there were still only a few more hits. The largest number of false alarms occurred when a transition was predicted to go from Grassland to Building but stayed at Grassland. These occurred around parks and golf courses where it would not make sense for an expansion to occur. There were two transitions that accounted for most of the misses in this model. One transition was Trees that were predicted to persist as that class but instead turned to Buildings; the other was Grassland that was also predicted to stay as Grassland but instead turned to Buildings. There were more misses of the latter. In general, the model is not complex enough to understand building permits, densification, or expansion of highways. Therefore, even if this model had a higher accuracy rate with more transitions, there still would be many pixels that would either be false alarms or misses due to the complexity of the study area.

The 2100 LULC (Figure 7), 2200 LULC (Figure 8) and 2300 LULC (Figure 9) rasters were calculated in TerrSet's Land Change Modeler. These LULCs for the North Shore are used to compare what the predicted land change will be when the SLR rises to a certain height based on the results of the MLP-MC. When these three LULCs are compared with the current LULC that Esri created for 2021, a one-to-one comparison cannot be used since Esri used Sentinel-2 bands (10m resolution) whereas the data that was created in TerrSet used Landsat-5 bands (30m resolution). When a mask of the areas that will be inundated is used, the pixels defer between the 2021 LULC and the future predicted LULCs in areas with 1m and 2m SLR showing a larger area with the 2021 LULC. By contrast, 4m SLR has a larger area with the predicted 2300 LULC. Moreover, these two LULC types use different classifications for each type of land cover, where beaches considered to be Water in the Esri LULC are Buildings or Roads in the TerraSet

LULCs. This heavily impacts the 1m SLR since the 2100 predicted LULC has 0% of pixels as part of the Waterbodies classification and 97% as part of the Buildings or Roads Classification; the 2021 LULC has 47.94% of pixels as the Water class and 48.71% as Built Areas. However, even with beach area as a large part of the 1m SLR, over the more than 275 years of predicted SLR for 4m to occur, there will not be a large increase in the class Buildings or Roads, most of which in the study area will not be inundated due to the rapid increase of elevation away from the coastline. For example, when the area of 4m SLR is used to compare the number of pixels that are part of the class Buildings or Roads between the two LULCs of 2100 and 2300, there only were 12 more pixels that were classified as Buildings or Roads in the 2300 LULC compared to the 2100 LULC. This demonstrates there is not a large increase of predicted built area in this inundated zone.

Table 8 Number and percentage of 30m resolution pixels per predicted future LULC class for each SLR height

| SLR (year) | Water | Trees | Buildings or Roads | Bare Earth | Grassland |
|---|---|---|---|---|---|
| 1m (2100) | 0 at 0% | 2 at 0.41% | 473 at 97.12% | 3 at 0.62% | 9 at 1.85% |
| 2m (2200) | 0 at 0% | 19 at 2.50% | 669 at 88.14% | 5 at 0.66% | 66 at 8.70% |
| 4m (2300) | 19 at 0.30% | 423 at 6.58% | 5543 at 86.23% | 135 at 2.10% | 308 at 4.79% |

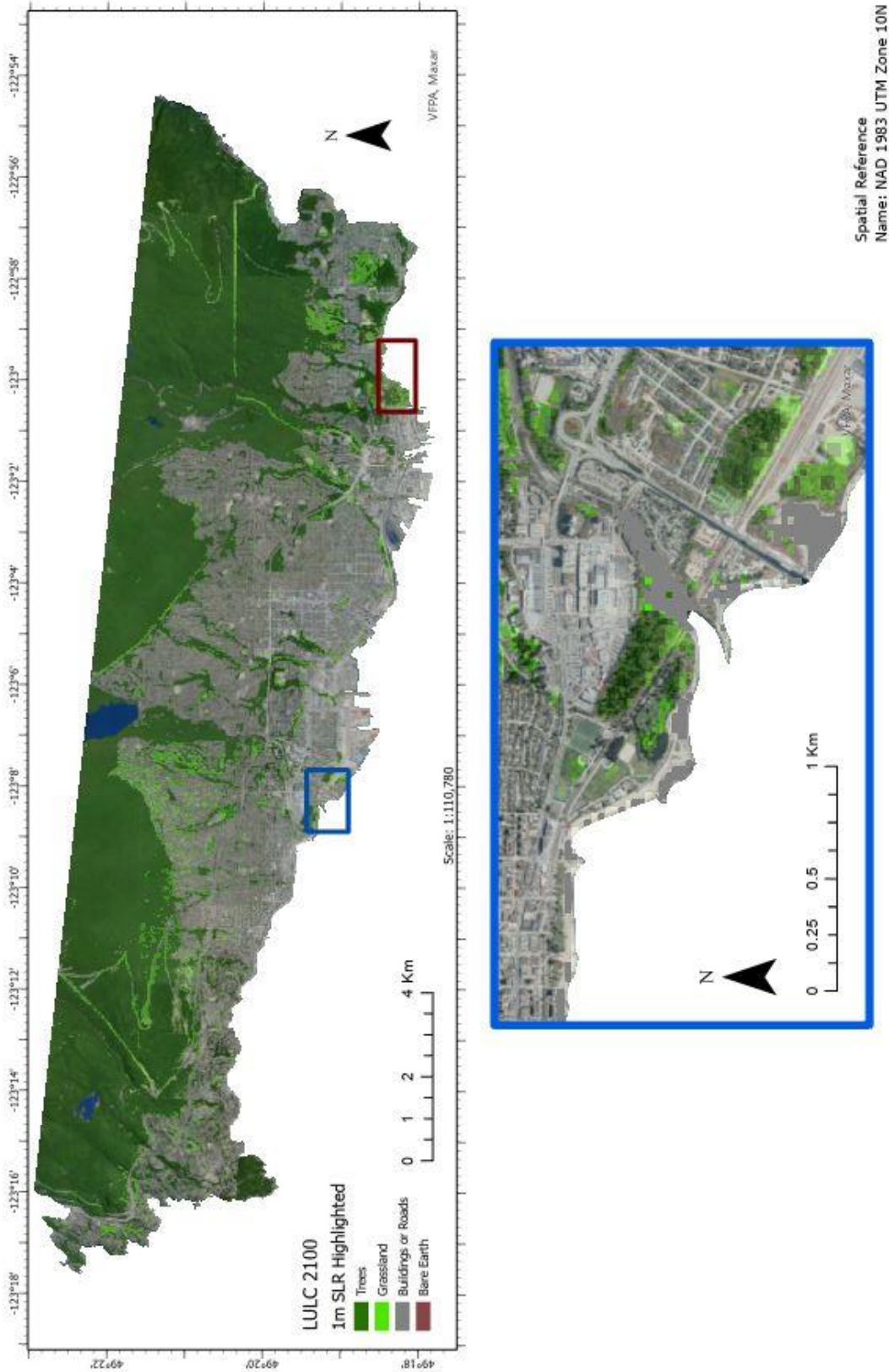

Figure 7 Predicted LULC for when 1m SLR is likely to occur in the study area

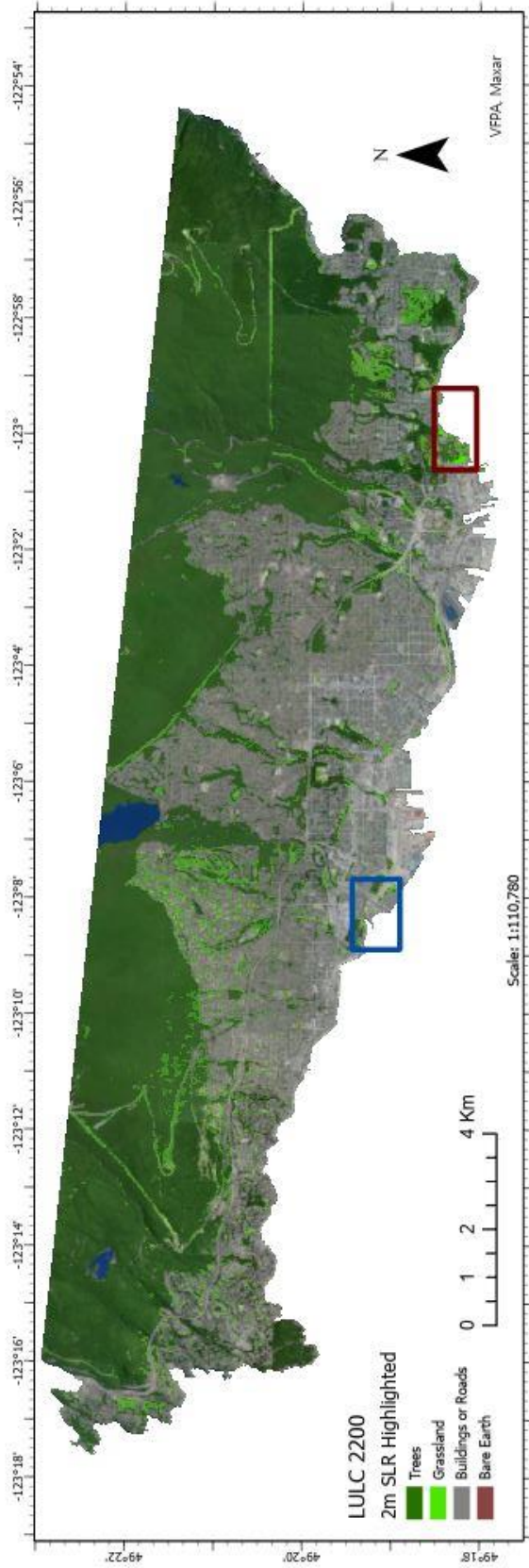
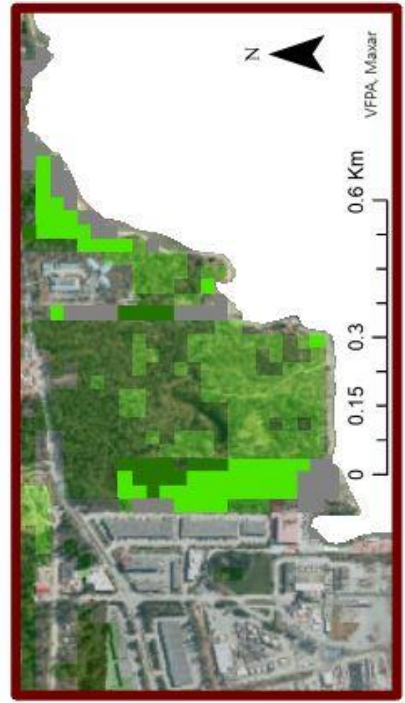
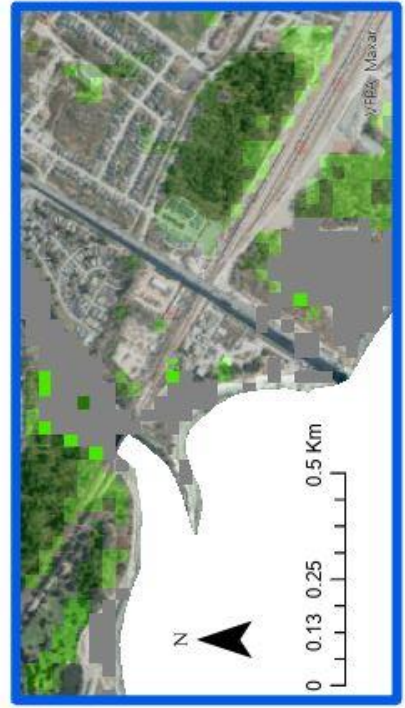

Figure 8 Predicted LULC for when 2m SLR is likely to occur in the study area

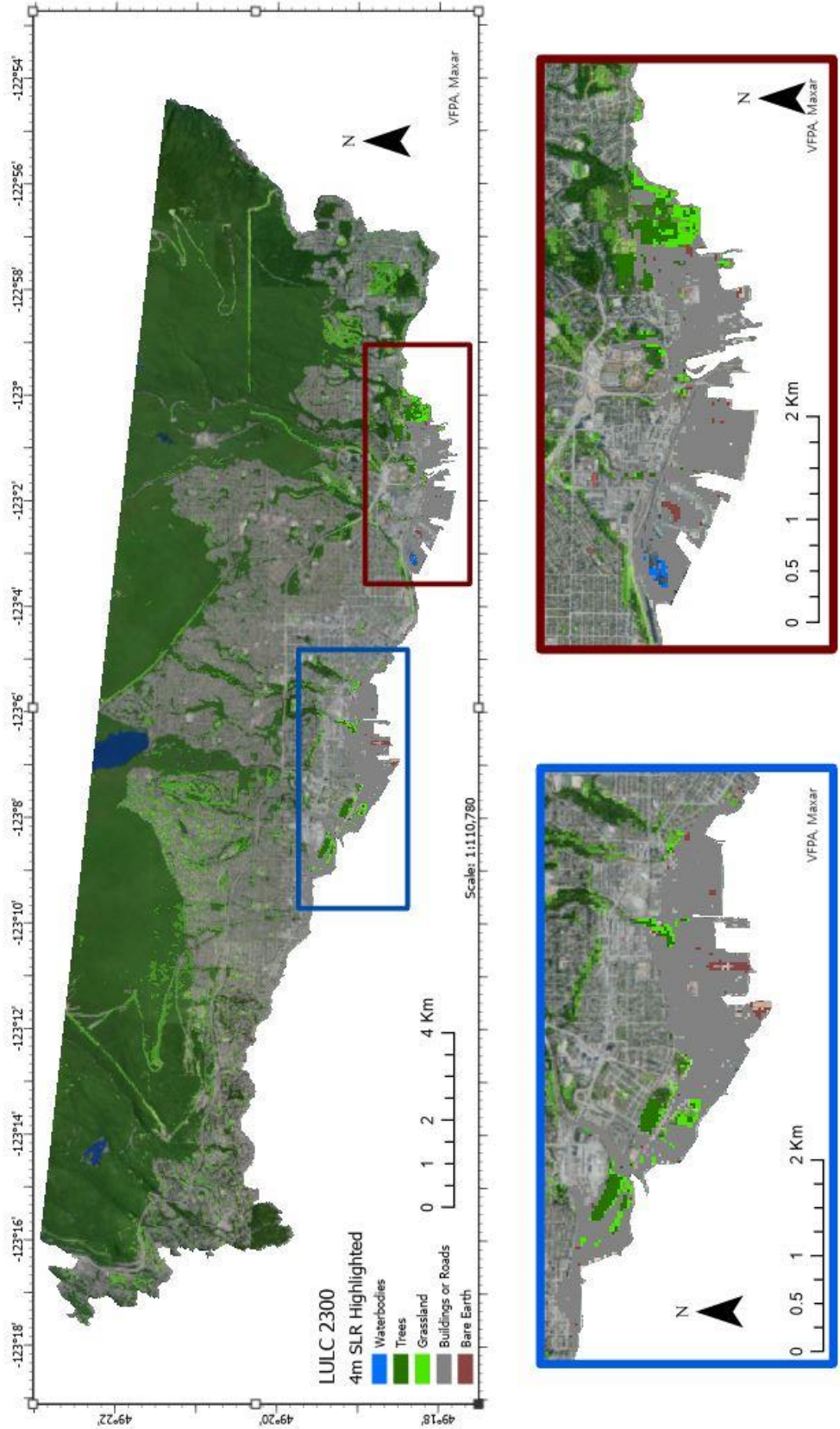

Figure 9 Predicted LULC for when 4m SLR is likely to occur in the study area

Even though this LULC land change process had a high accuracy rate, the validation between LULC prediction and actual LULC displayed numerous misses. This might be due to the unusual land use situation of the North Shore of Vancouver where both its unusual topography and population growth produce quite an unpredictable land use direction. This region, surrounded by coastal waters and steep mountains with minimal suburban sprawl, has several golf courses and numerous parks that probably will not transition into buildings. While it has some residential growth up the sides of the three mountains, this is quite limited. Instead, the region mainly develops in its densely populated urban areas by replacing single-family houses with condominiums and apartments. All these factors make it difficult to predict future land use.

### 4.5. Results from Online Map

A layer for buildings of the study area was created so that a 3D visualization would not only include the DTM and SLR layers, but also individual buildings affected by inundation. Ships were sometimes mistaken for buildings, this error did not affect the final product since the current study area is land-based, only analysing buildings that would directly be affected by flooding. This included areas that barely touched any part of the 4m inundation zones. Although overall 1000 polygons were created, some of these polygons were parts of one building. In general, the heights of most of the buildings were between 5m and 20m with a few apartment buildings being over 40m. Only several buildings would be completely inundated if SLR of 4m does occur. Showing individual buildings to the public that will be affected by SLR demonstrates the scale of businesses and homes that will be destroyed in the future due to Climate Change.

The impacts of inundation due to Climate Change were created using: 3D Local Scene of the layers of SLR, the 1m resolution DTM, buildings that will be partially inundated by SLR as well as a satellite image of the study area. Each of the SLR layers and buildings were extruded to show the height of inundation that could occur and how it will affect the buildings situated in this zone. Besides the online maps this 3D visualization allows the public to view major hotspots for flooding by using the interactive map to focus on specific areas.

Popups of important locations were created on the online map (Figure 10) for this study to demonstrate to the public the number of important features or sites that will be affected by SLR. Nine popups were created which include a nature reserve, a ferry port, international shipping ports, as well as one of the largest shopping malls in Canada. Other than Horseshoe Bay which houses the ferry port and town in the far western section of this region, all of the other popups are part of the two southern areas of the Narrows that will be severely affected by SLR. Seven of the major features will be only partially or fully inundated at 4m SLR whereas Ambleside Park will be fully flooded at 3m SLR. Even though most of the flooding will occur in Horseshoe Bay at 2m SLR, only a small portion will be inundated mostly on the west coast, not significantly affecting the ferry terminals. The interactive map popups which include a non-copyrighted image, a text description and a link to the respective websites demonstrate to the public well known or important local environmental or economic locations that will be inundated due to SLR. Raising awareness by showing to the public future locations at risk of inundation can create a personal reaction to the impacts of Climate Change (Retchless, 2018). This interactive map was then turned into an Esri Story Map. The link to the Story Map is: https://storymaps.arcgis.com/stories/1d1ee6d911b84e66bb76714964cef6af. Besides the addition of this map, a 3D flyover video, a small blurb of the location as well as a description of the data in the interactive map was included.

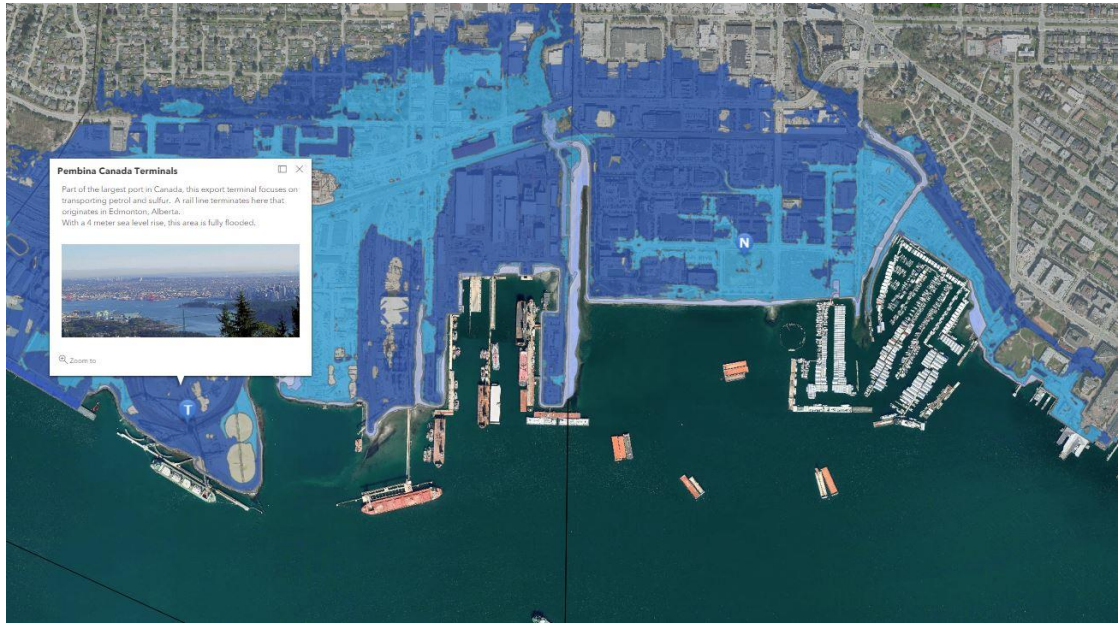

Figure 10 Section of the interactive map showing part of the First Narrows with a popup of Pembina Canada Terminals being open.

## 5. Discussion

Using the results of the data a more precise, publicly accessible interactive online map was successfully created that allows the public to explore the impact of SLR due to climate change with its direct impact to key well-known places of the North Shore of Vancouver, Canada. Retchless (2018) discussed how these devastating SLR predictions are difficult for the public to fully understand and noted the importance of visualizing which local areas will be inundated since the height of several metres is difficult for the public to conceptualize. Through the interactive online map that was created the local population of the North Shore of Vancouver will be able to observe that most of the inundation that will occur in this study area is between 3m and 4m SLR. Moreover, through present and predicted future LULCs most of the areas that will be inundated will affect built areas such as infrastructure, roads, buildings and or other unnatural areas. Through using Deep Learning to find the building footprint of the region, this study found that hundreds of buildings will be affected due to the predicted inundation, but this mainly occurs at over 2m SLR.

Even though results of this data indicate that most of the study area does not inundate until 3m SLR, the municipalities of the study area will be in a dire situation in the future without actions taken to mitigate climate change globally. This is due to the number of important industries that would be partially or fully inundated if 4m SLR occurs in this region. Besides the two shipping terminals in this study area that were used as popups in the Story Map there are several other terminals that comprise the Port of Vancouver which will be fully flood. Moreover, these terminals in the North Shore that transport mostly raw resources such as petroleum coke and sulphur would pollute Burrard Inlet as is also the case for the waste of The North Shore Waste and Recycling Centre. Even if all these resources and waste were moved, there still would trace amounts of substances to pollute the inlet and surrounding waterways. Other than its rail line, the G3 Terminal will be fully flooded at 4m.  This terminal exports most of Canada's grain, not transported to the United States, to Asia and the rest of the world. If this area were flooded, it will not only cause severe economic repercussions to the North Shore but also to the rest of Canada. In addition, most of the rail lines which are close to the coast are used by freight trains to transport grain and resources to the terminals. These lines could be moved to higher elevation as a precaution, but they would still have to be connected somehow to the Port of Vancouver. Even though most of the industrial and economic transportation will be affected by 3m and 4m SLR in this region, this is not the case for road transportation since the two bridges as well as the major highway of this region would not be significantly inundated.

It is also important to discuss our results in light of the key coastal geovisualisation studies. As the first such perspective article on SLR-related geovisualisation, Richards (2015) offered some useful insights on employing effective research communication through risk-based interactive geovisualisation technologies as productive usability of online, participatory technologies that promote citizen engagement in science. Risk Finder tool (https://sealevel.climatecentral.org/about/) launched in 2013 by non-profit climate communication and research group Climate Central is the most noticeable initiative on enabling SLR geovisualisation for USA coastal communities using high-resolution LiDAR datasets. Similar LiDAR-based geovisualisation was lacking for our study area, which we have achieved through this work. Newell and Canessa (2017) came up with a place-based concept on developing geovisualisations for coastal planning, as they rightly acknowledged how different user groups relate to coastal environments as "places" of values and meanings, rather than simply 'spaces' which a traditional spatial analysis in GIS focusses at. This concept of "places" receives relevance if the study is performed at a local or regional scale, an approach that we have adopted in our work. Wherever possible, we have discussed our results in terms of local areas and buildings of relevance, offering a context to the local user groups of our interactive map. In Canadian context, Minano et al. (2018) developed a Geoweb tool called AdaptNS for supporting local climate change adaptation efforts in coastal communities of Nova Scotia. AdaptNS as a web-based geovisualization tool displays interactive inundation maps generated using LiDAR data, local climate change projections of SLR, and storm surge impacts between the years 2000 and 2100 (Minano et al., 2018). In our paper, we have added similar geovisualisation for another long Canadian coastline and the usability of LiDAR data for making the geovisualisation reliable and effective was displayed.

## 6. Conclusion

The aim of this study is to produce an interactive online map that is not only accessible to the public, but also allows them to interact with the newest high-resolution data and techniques possible within the scope of this study. Besides the interactive online map, 3D visualizations of 1m, 2m, 3m, and 4m SLR layers and a newly created 3D extruded building layer were used to create a 3D flyover animation to also engage the public. This study effectively meets the three primary objectives described in the Aims and Objectives section of the Introduction by utilizing an interactive map (3) and land use data (2) to showcase the impact of sea level rise (1) on the North Shore of Vancouver. This project demonstrates how 4.3% of the study area in the North Shore of Vancouver due to SLR could be inundated with major industry, protected areas and commercial sites being destroyed. Through the data that was created, the public local community can observe the extent and severity that might occur if mitigation is not implemented to curb Climate Change. The interactive map setup encourages user interaction, exploration, and reflection of SLR impacts on their communities.

While this 4.3% SLR inundation will flood a great deal of the coastline of the North Shore of Vancouver, it should be noted that most of the major flooding that might occur would disrupt commercial, environmental, and industrial areas only at 3m and 4m SLR. This level of SLR is some of the worst-case scenarios and will not likely occur until the end of the 2200s or early 2300s. However, to avoid such scenarios, mitigation efforts to curb the effects of Climate Change need to be taken and not delayed so that such a future is not a reality. Throughout this paper, we have taken steps to provide a detailed and easily understandable description of methods, so that it becomes easier to reproduce these steps for other regions, aiding SLR geovisualisation for a wider coastal community. The next step for us is to work towards effectively combining geo and demographic datasets to develop risk maps for a wider study area, resulting in a publicly-accessible interactive GIS capable of sourcing local datasets on SLR and social variables (e.g., demography and property values).